# Diagnosis of Parkinson's Disease Based on Voice Signals Using SHAP and Hard Voting Ensemble Method

Paria Ghaheri[1], Hamid Nasiri[2,*], Ahmadreza Shateri[1], Arman Homafar[1]

[1] Electrical and Computer Engineering Department, Semnan University, Semnan, Iran
[2] Department of Computer Engineering, Amirkabir University of Technology (Tehran Polytechnic), Tehran, Iran.

Correspondence should be addressed to Hamid Nasiri; h.nasiri@aut.ac.ir

## Abstract

**Background and Objective:** Parkinson's disease (PD) is the second most common progressive neurological condition after Alzheimer's, characterized by motor and non-motor symptoms. Developing a method to diagnose the condition in its beginning phases is essential because of the significant number of individuals afflicting with this illness. PD is typically identified using motor symptoms or other Neuroimaging techniques, such as DATSCAN and SPECT. These methods are expensive, time-consuming, and unavailable to the general public; furthermore, they are not very accurate. These constraints encouraged us to develop a novel technique using SHAP and Hard Voting Ensemble Method based on voice signals.

**Methods:** In this article, we used Pearson Correlation Coefficients to understand the relationship between input features and the output, and finally, input features with high correlation were selected. These selected features were classified by the Extreme Gradient Boosting (XGBoost), Light Gradient Boosting Machine (LightGBM), Gradient Boosting, and Bagging. Moreover, the Hard Voting Ensemble Method was determined based on the performance of the four classifiers. At the final stage, we proposed Shapley Additive exPlanations (SHAP) to rank the features according to their significance in diagnosing Parkinson's disease.

**Results and Conclusion:** The proposed method achieved 85.42% accuracy, 84.94% F1-score, 86.77% precision, 87.62% specificity, and 83.20% sensitivity. The study's findings demonstrated that the proposed method outperformed state-of-the-art approaches and can assist physicians in diagnosing Parkinson's cases.

## 1. Introduction

Parkinson's disease (PD) is the most widespread movement condition and the second most prevalent neurodegenerative illness after Alzheimer's [1]. The etiology of this disease is the slow degeneration of dopaminergic neurons in the midbrain, which results in various motor and non-motor problems [2]. The most noticeable symptoms of Parkinson's disease are tremors, rigidity, bradykinesia, postural instability, slow movement, hyposmia, sleep difficulties, and changes in voice tone [3]. Patients with PD typically begin experiencing symptoms at age 58; however, symptoms can begin as early as 40 in some cases [4]. Over 8.5 million people worldwide were estimated to have Parkinson's disease in 2019. According to current estimates, Parkinson's disease caused 329,000 deaths and 5.8 million years of disability-adjusted life in 2019 (an increase of 81% from 2000) [5]. The Global Burden of Disease Study projects that 13 million PD cases will be by 2040 [6].

Although non-motor symptoms are present in many individuals before the start of PD, they lack specificity, are challenging to evaluate, and vary from patient to patient [7]. Consequently, in most cases, Parkinson's disease is initially diagnosed based on motor symptoms. There are a total of three steps necessary to diagnose the condition. The physician often determines the first stage based on the patient's complaints and the neurological examination that follows the disease history [8]. Following the confirmation of symptoms, the next step is drug therapy, in which the patient with probable Parkinson's disease is treated with dopamine. If the condition improves, there is a significant likelihood that the individual has the disease [9]. The lack of a solid and consistent clinical response to the dopaminergic medication has driven the search for disease biomarkers, including imaging and laboratory-based techniques, known as the third stage [10]. Imaging diagnostic procedures can detect issues in the brain. However, they are invasive, expensive, or insufficiently specific, such as single photon emission computed tomography (SPECT), DaTSCAN or nuclear magnetic resonance with diffusion tensor imaging (NMR-DTI) [11]. On the other hand, not everyone may have access to laboratory tests, particularly in underdeveloped nations [9]. As a result, it is critical to develop an early diagnostic approach that is simple, quick, accurate, and open to the general public.

Up to 89 % of Parkinson's patients experience speech difficulties [12]. Common perceptual symptoms include reduced loudness (hypophonia), pitch change (monotone), breathy and harsh voice quality, and inaccurate articulation [13]. Therefore, recording the patient's voice is a beneficial diagnostic technique since people with Parkinson's disease have distinctive vocal traits. Machine Learning (ML), a subfield of artificial intelligence, is gaining traction in the medical field because of its potential to improve the course of treatment for patients with chronic diseases [14]. Consequently, applying machine learning algorithms to a collection of speech recordings to detect Parkinson's disease properly would be a feasible screening step before seeing a clinician [15].

This study utilizes a feature selection method based on Pearson Correlation Coefficients (PCCs) to determine the optimal number of features for classification. After extracting features, they are given as input to the hard voting ensemble method. In the final phase, a novel explainable artificial intelligence (XAI) technique, i.e., Shapley Additive exPlanations (SHAP), is employed to gain a more comprehensive understanding of the effect of the features on the output.

## 2. Related Works

As the prevalence of Parkinson's disease rises, numerous researchers have demonstrated a keen interest in developing a diagnostic method. Using machine learning has significantly advanced this strategy. Karabayir et al. [16] evaluated the data using several ML methods, including Extreme Gradient Boosting (XGBoost), Light Gradient Boosting Machine (LightGBM), Random Forest (RF), Support Vector Machines (SVM), K-nearest Neighbors (KNN), Least Absolute Shrinkage and Selection Operator Regression (LASSO), as well as Logistic Regression (LR). In addition, they used a variable significance analysis to identify critical characteristics for diagnosing patients with PD, achieving an accuracy of 84.10%. Magesh et al. [17] developed a ML model capable of classifying each DaTSCAN as having Parkinson's disease or not. Using Local Interpretable Model-Agnostic Explainer (LIME) techniques, visual indicators were employed to convey the reasoning behind the prediction. Transfer learning was utilized to train DaTSCANs using the Parkinson's Progression Markers Initiative database on a Convolutional Neural Network (CNN).

The approach proposed by Sajal et al. [18] takes in rest tremor and vowel phonation data from smartphones equipped with accelerometer and voice recorder sensors. They employed the Maximum Relevance Minimal Redundancy (MRMR) algorithm for feature selection, a technique based on information theory. After selecting features, they investigated and optimized different classifiers, including KNN, SVM, and Naive Bayes. In the end, KNN demonstrated the highest accuracy and the best performance. Mittal and Sharma [19] proposed a classification method for Parkinson's illness that combines data partitioning with the algorithm for feature selection and Principal Component Analysis (PCA). They also used three different classifiers to classify all data partitions, including the weighted KNN, LR, and Medium Gaussian Kernel SVM (MGSVM).

Karaman et al. [8] created CNNs for automatic PD identification using voice signals obtained from biomarkers. SqueezeNet, ResNet101, and DenseNet161 architectures were retrained and assessed to find which architecture can reliably categorize frequency-time information. Mohammadi et al. [20] proposed autoencoders as efficient feature extractors. Stacking a combination of SVM, XGBoost, Multilayer Perceptron (MLP), and SVM, they discriminated against PD patients from usual ones. Also, a voting mechanism was employed to improve each classifier's predictions.

Pramanik et al. [21] created a model based on the vocal fold, time-frequency, and baseline features of people with PD. They began by putting the vocal characteristics through a series of tests, including correlation, the fisher score, and a mutual information-based feature selection scheme. The assessed features were progressively sent to multiple classifiers, with Naive Bayes emerging as the top classifier for the proposed model. Rahman et al. [22] assessed the difficulty of PD identification based on various voice signal forms. A signal processing algorithm, i.e., MFCC, was used to extract numerical features from the speech phonations. Through the use of the Linear Discriminant Analysis (LDA) model, the dimensionality of the retrieved MFFC features was decreased. Different machine learning models were developed during the final phase.

Lamba et al. [23] proposed a hybrid PD detection system based on speech signals. Three feature selection approaches were used, including mutual information gain, extra tree, and genetic algorithm, as well as three classifiers, including Naive bayes, KNN, and RF. The imbalance of the dataset was fixed by the Synthetic Minority Oversampling Technique (SMOTE). Nilashi et al. [24] employed unsupervised and supervised learning approaches to apply UPDRS prediction to identify PD. The study showed that deploying Expectation-maximization with Support vector regression (SVR) ensembles outperformed Decision Trees (DT), Neuro-fuzzy, and SVR paired with other clustering techniques in predicting Motor-UPDRS and Total-UPDRS.

## 3. Materials and Methods

The proposed method employs the Pearson Correlation Coefficients, Hard Voting Ensemble Method, SHAP, and different classification algorithms, including XGBoost, LightGBM, Gradient Boosting, and Bagging, which will be discussed in this section.

### 3.1. Dataset

We used the "Parkinson Dataset with Replicated Acoustic Features" given to the Machine Learning repository at the University of California, Irvine, in April 2019 by Naranjo et al. [25]. In this dataset, characteristics are gathered by examining the sound recordings of 80 individuals (40: healthy, 40: PD). Three repetitions of a five-second-long phonation of the vowel /a/ were performed. The sampling rate for digital recordings was 44.1 kHz, and the sample size was 16 bits. The collected features were separated into various groups based on whether or not they share similar formulations. This segmentation yielded nine groups, four of which had a single characteristic (Table 1).

### 3.2. Shapley Additive Explanations

SHAP refers to "Shapley Additive exPlanations," a ML technique that explains model predictions and provides interpretability for ML models [26]. This strategy entails retraining the model on all subsets of features $S \subseteq F$, where F is the complete collection of features. It assigns a significant value to each feature, indicating that feature's influence on model prediction. To extract the impacts of factor $i$ two models are trained. The first model $f_{S \cup \{i\}}(x_{S \cup \{i\}})$ is trained with factor $i$ while the other one $f_S(x_S)$ is trained without it. Then the predictions of these two models are compared to the present input $f_{S \cup \{i\}}(x_{S \cup \{i\}}) - f_S(x_S)$, where $x_S$ represents the values of the input features in the set $S$ [27], [28]. According to cooperative game theory, Shapley value for the model can be computed as follows:

$$\phi_i = \sum_{S \subseteq F \setminus \{i\}} \frac{|S|! \, (|F| - |S| - 1)!}{|F|!} [f_{S \cup \{i\}}(x_{S \cup \{i\}}) - f_S(x_S)] \qquad (1)$$

The SHAP technique approximates the Shapley value by performing Shapley sampling or Shapley quantitative influence to estimate $\phi_i$ from $2^{|F|}$ differences. According to this interpretation, the SHAP value for each variable may be established, and input parameters can be sorted according to their $\phi_i$ [29]. The SHAP value is the sole explanation approach supported by a sound theory and the most locally precise and consistent feature contribution value attainable. The most innovative technique for approximating SHAP values is Tree SHAP, which accurately computes Shapley explanations. It uses DT structures to separate each input's contribution in a DT or ensemble DT model [30]. A trained model is required as an input for the Tree SHAP technique, and for this study, our final model will stand in for the trained model.

### 3.3. Gradient Boosting Decision Tree

Gradient boosting is the most advantageous ML technique for classification and regression problems. It builds a prediction framework by combining weak prediction frameworks, primarily DTs [31]. The Gradient boosting decision tree (GBDT) technique uses Gradient boosting to extend and improve the classification and regression tree models. DTs are constructed iteratively by the GBDT algorithm. In each iteration, a DT is trained using the residuals from the preceding tree. Ultimately, the output is determined by accumulating the classified results of each tree [32].

Table 1: Description of the Parkinson Dataset with Replicated Acoustic Features

| Name of the group | #Features | Name of the features | Min | Max | Average | Standard Deviation |
|---|---|---|---|---|---|---|
| Pitch local perturbation measures | 4 | Relative jitter, Absolute jitter, Relative average perturbation (RAP), Pitch perturbation quotient (PPQ) | 0.00 | 6.84 | 0.15 | 0.13 |
| Amplitude perturbation | 5 | Local shimmer, Shimmer in dB, 3-point Amplitude Perturbation Quotient (APQ3), 5-point Amplitude Perturbation Quotient (APQ5), 11-point Amplitude Perturbation Quotient (APQ11) | 0.00 | 1.75 | 0.09 | 0.02 |
| Harmonic-to-noise measures | 5 | HNR05 [0–500 Hz], HNR15 [0–1500 Hz], HNR25 [0–2500 Hz], HNR35 [0–3500 Hz], HNR38 [0–3800 Hz] | 22.22 | 129.99 | 71.78 | 342.05 |
| Mel frequency cepstral coefficient-based | 13 | MFCC0, MFCC1, . . ., MFCC12 | 0.57 | 2.07 | 1.34 | 0.05 |
| Derivatives of Mel frequency cepstral coefficients | 13 | Delta0, Delta1, . . ., Delta12 | 0.62 | 2.04 | 1.34 | 0.05 |
| Recurrence Period Density Entropy (RPDE) | 1 | - | 0.16 | 0.54 | 0.31 | 0.00 |
| Detrended Fluctuation Analysis (DFA) | 1 | - | 0.41 | 0.78 | 0.61 | 0.01 |
| Pitch Period Entropy (PPE) | 1 | - | 0.00 | 0.91 | 0.27 | 0.05 |
| Glottal-to-Noise Excitation Ratio (GNE) | 1 | - | 0.85 | 0.99 | 0.92 | 0.00 |

## 3.4. Extreme Gradient Boosting

Extreme Gradient Boosting (XGBoost) is a sophisticated supervised method presented by Chen and Guestrin inside the context of tree boosting, which boasts rapid, precise, efficient operations and strong generalizability [33], [34]. Data scientists frequently use it to produce state-of-the-art outcomes on various ML issues [35]. In fact, XGBoost is an enhanced version of the GBDT method [36], consisting of many DTs, and is commonly employed in classification and regression. However, XGBoost is distinct from GBDT in some ways. Firstly, XGBoost augments the loss function using a second-order Taylor expansion, whereas the GBDT method utilizes a first-order Taylor expansion [37], [38]. Second, the objective function uses normalization to prevent overfitting and simplify the approach [39], [40]. In other words, parallel and distributed computing contributes to a faster learning process [29]. The prediction functions for time step *t* are as follows:

$$f_i^{(t)} = \sum_{k=1}^{t} f_k(x_i) = f_i^{t-1} + f_t(x_i) \tag{2}$$

The depth and the number of trees are crucial factors for the XGBoost algorithm. The challenge of locating the optimal algorithm was recast as a new classifier that can lower the loss function, with the target loss function represented by (3) [41].

$$Obj^{(t)} = \sum_{k=1}^{n} l(\bar{y}_i, y_i) + \sum_{k=1}^{t} \Omega(f_i) \tag{3}$$

where *l* is the loss function, *n* is the observation number, $\Omega$ is the regularization term, which is expressed as:

$$\Omega(f) = \gamma T^2 + \frac{1}{2}\lambda\|w\|^2 \tag{4}$$

where $\omega$ is the score vector, $\lambda$ is the regularization parameter, and $\gamma$ is the mini loss [42]. Several parameters must be modified before utilizing the model to optimize its performance and prevent excessive or insufficient fitting [43], [44].

## 3.5. Light GBM

Microsoft introduced the Light Gradient Boosting Machine (LightGBM) method, which is now a popular open-source, distributed, high-performance Gradient-Boosting framework [45]. LightGBM is constructed using DT and histogram methods, which make data segmentation easier. LightGBM comprises the algorithms Gradient-based One-Side Sampling (GOSS) and Exclusive Feature Bundling (EFB) [46]. GOSS is a type of down-sampling strategy. During the model training phase, cases with more substantial gradients contribute more to the information gain. Therefore, GOSS retains instances with significant gradients and randomly removes ones with slight gradients when down-sampling the data [47], [48]. Also, the EFB method seeks to reduce the number of features by grouping together several exclusive qualities, which can significantly reduce the number of needless calculations for zero feature values [49]. LightGBM, in contrast to typical level-wise methods, employs a leaf-wise approach with depth limiting to boost accuracy and prevent overfitting [50].

### 3.6. Bagging

Bagging is an ensemble approach developed by Leo Breiman in 1994 [51]. It is defined by the production of multiple samples using bootstrap refitting from the same data set so that many different trees for the exact predictor can be constructed and used to develop an aggregate prediction [52]. Bagging tries to reduce variance while preserving the bias of a DT, to minimize overfitting, and to enhance the precision and consistency of ML algorithms [53].

### 3.7. Proposed Method

In this section, the proposed approach for the diagnosis of PD is presented. In this approach, first, we determine the PCCs for each acoustic characteristic. We take into consideration a threshold, and the features whose absolute correlation value with the output is more than the threshold (i.e., 0.47) are the ones that are selected. Second, four classifiers, namely XGBoost, LightGBM, Gradient boosting, and Bagging, utilize the 32 selected features as input and classify the patients. Third, we use the Weighted Hard Voting Ensemble of the four classifiers in which the models' majority vote determines the classification based on different weights. Finally, we apply SHAP to the model so that it can assess the importance of features. Fig. 1 depicts the general framework of the proposed method.

## 4. Results and Discussion

The proposed method was evaluated on Parkinson's Dataset with Replicated Acoustic Features. Several performance metrics, including precision, sensitivity, specificity, $F_1$-score, and accuracy, were used to assess multiple ML models using the proposed methodology. According to the count of true positive (TP), false positive (FP), true negative (TN), and false negative (FN), the following equations can be used to calculate mentioned evaluation metrics [54]:

$$Precision = \frac{TP}{TP + FP} \qquad (5)$$

$$Sensitivity = \frac{TP}{TP + FN} \qquad (6)$$

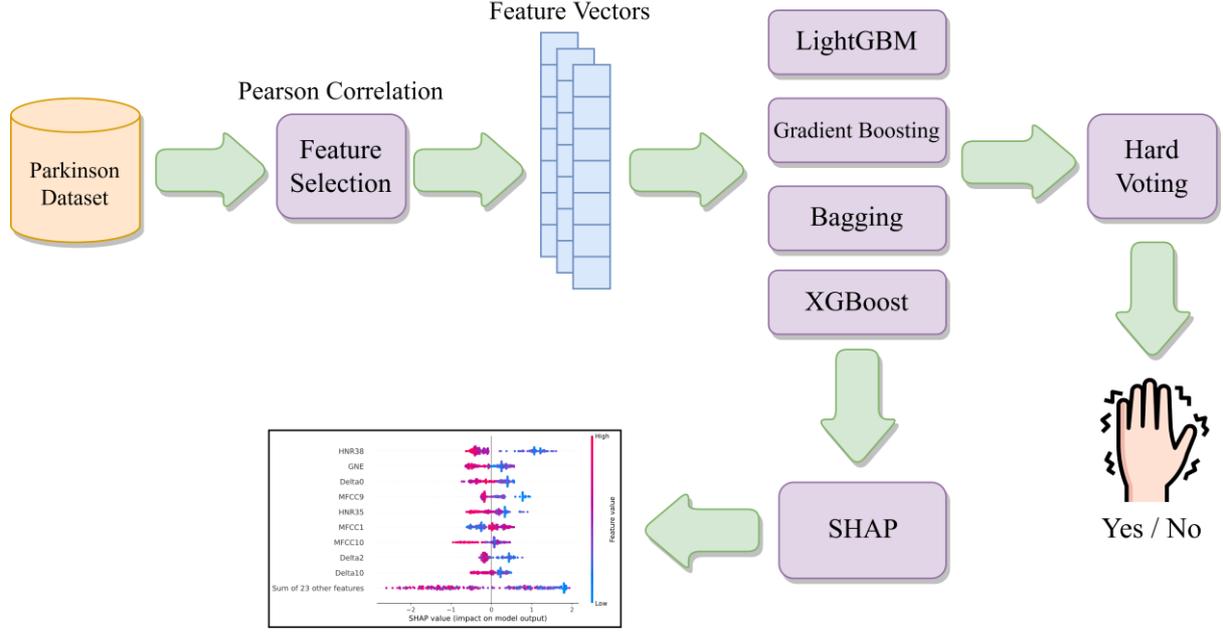

Figure 1: The architecture of the proposed model

$$Specificity = \frac{TN}{TN + FP} \quad (7)$$

$$F_1 - Score = \frac{2 \times Sensitivity \times precision}{Sensitivity + precision} \quad (8)$$

$$Accuracy = \frac{TP + TN}{TP + FN + FP + TN} \quad (9)$$

In the initial phase of this methodology, to reduce classification time and improve classifier performance, the PCCs were computed for the entire dataset (Fig. 2). This method was utilized to comprehend the relationship between each feature and the output. The PCCs demonstrated that the four features, MFCC10, Delta11, HNR35, and HNR38, correlated highly with the output. In contrast, DFA, RPDE, Jitter_abs, and Gender features had a low correlation. We consider a threshold, i.e., 0.47, for evaluating highly correlated features. Consequently, 32 features that correlated above that threshold are considered for the following phase.

In the second phase, four candidate classifiers were applied to the dataset of 32 features acquired in the preceding step: XGBoost, LightGBM, Gradient boosting, and Bagging. When configuring XGBoost's hyperparameters, we discovered that it offered great flexibility. Consequently, the XGBoost parameters were acquired by trial and error, as shown in Table 2. On the other hand, using a histogram-based approach, LightGBM achieved a faster training speed and greater efficiency. Also, Bagging had the advantage of allowing a group of weak learners to outperform a single strong learner by combining their efforts. It assisted in reducing variance, hence preventing the overfitting of the model during the approach.

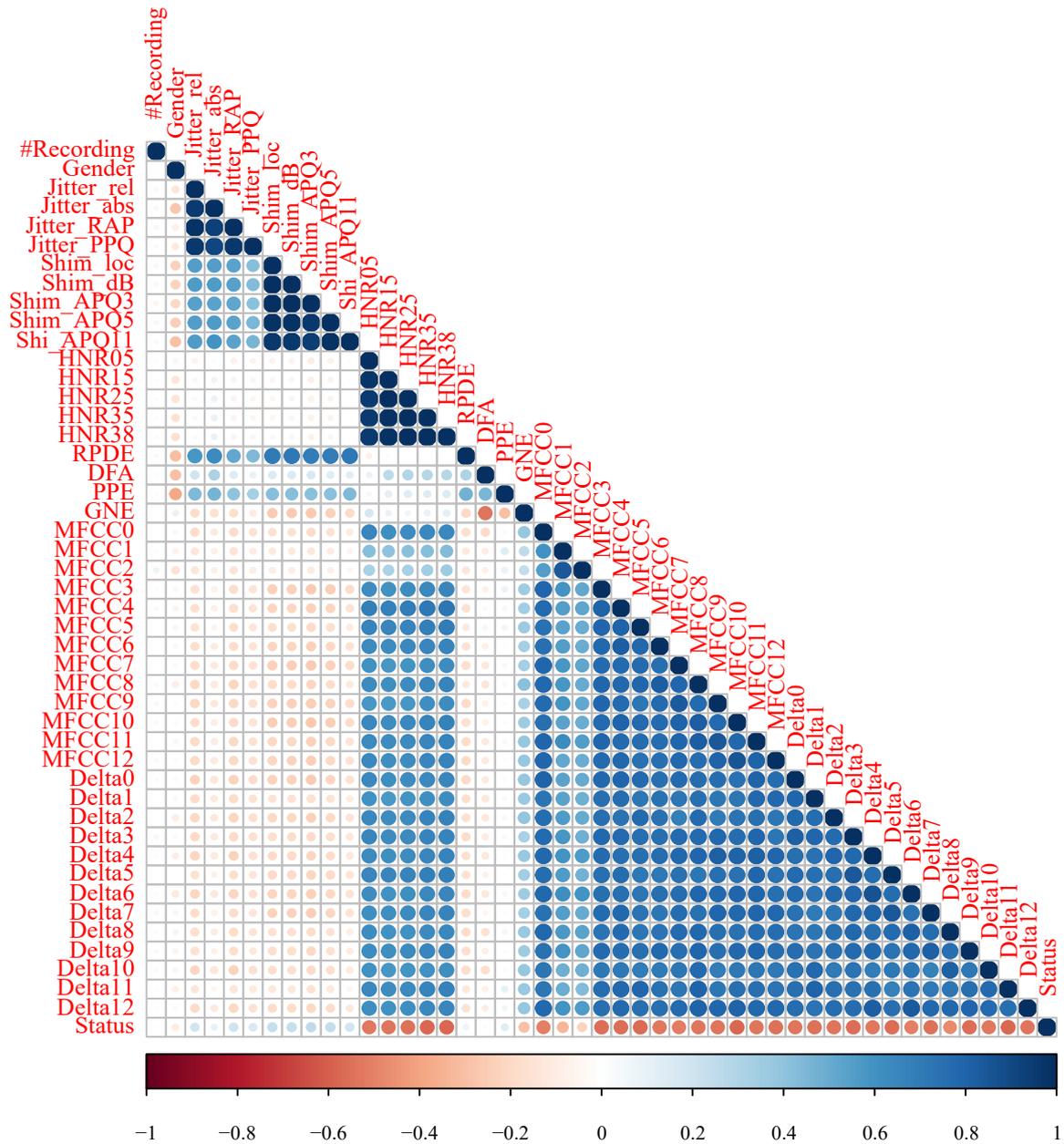

Figure 2: Pearson Correlation Coefficients for Parkinson's Dataset

Table 2: The XGBoost parameter setting

| Parameter | Value |
| --- | --- |
| Base learner | Gradient boosted tree |
| Tree construction learner | Exact greedy |
| Number of gradient boosted trees | 94 |
| Learning rate ($\eta$) | 0.001 |
| Lagrange multiplier ($\gamma$) | 0 |
| Maximum depth of trees | 6 |

Given that four classifier candidates performed reasonably well, we decided to use a Hard Voting Ensemble Method in the proposed strategy to improve its prediction accuracy further. To utilize the advantageous characteristics of each classifier to enhance accuracy, the weighting was set depending on each classifier's performance. Consequently, XGBoost, LightGBM, Gradient boosting, and Bagging acquired weights of 13, 6, 6, and 1, respectively. Finally, the majority vote of the classifiers determined the final prediction.

To further strengthen this investigation and demonstrate the proposed method's robustness, we employed 4-fold cross-validation. Cross-validation is a methodology for resampling a dataset to evaluate machine learning algorithms using limited data samples [55]. The 4-fold cross-validation technique divides the dataset into four samples, analyzes the method for each sample, and estimates the dataset's average accuracy.

Table 3 presents the comparison results of the proposed method with other classifiers in terms of accuracy, $F_1$-Score, precision, specificity, and sensitivity. As can be seen, our proposed method outperformed other classifiers with an accuracy of 85.42%, $F_1$-Score of 84.94%, precision of 86.77%, specificity of 87.62%, and sensitivity of 83.20%.

Table 3: Comparison of the proposed method with other classifiers

| Performance metrics (%) | Model | Folds-1 | Folds-2 | Folds-3 | Folds-4 | Average |
|---|---|---|---|---|---|---|
| Sensitivity | XGBoost | 77.77 | 81.81 | **87.50** | 85.71 | **83.20** |
| | LightGBM | **81.48** | **84.84** | 78.12 | 85.71 | 82.54 |
| | Gradient Boosting | 77.77 | 66.66 | **87.50** | 89.28 | 80.30 |
| | Bagging | 77.77 | 75.75 | 71.87 | 82.14 | 76.88 |
| | Proposed method | 77.78 | 81.82 | **87.50** | 85.71 | **83.20** |
| Specificity | XGBoost | 81.81 | 81.48 | 92.85 | **90.62** | 86.69 |
| | LightGBM | 78.78 | 85.18 | 92.85 | 81.25 | 84.52 |
| | Gradient Boosting | 81.81 | **88.88** | 85.71 | 87.50 | 85.98 |
| | Bagging | 75.75 | 85.18 | 92.85 | 87.50 | 85.32 |
| | Proposed method | **81.82** | 85.19 | **92.86** | 90.62 | **87.62** |
| Precision | XGBoost | 77.77 | 84.37 | **93.33** | 88.88 | 86.09 |
| | LightGBM | 75.86 | 87.50 | 92.59 | 80.00 | 83.98 |
| | Gradient Boosting | 77.77 | **88.00** | 87.50 | 86.20 | 84.87 |
| | Bagging | 72.41 | 86.20 | 92.00 | 85.18 | 83.95 |
| | Proposed method | **77.78** | 87.10 | **93.33** | 88.89 | **86.77** |
| $F_1$-Score | XGBoost | 77.77 | 83.07 | **90.32** | 87.27 | 84.61 |
| | LightGBM | **78.57** | **86.15** | 84.74 | 82.75 | 83.05 |
| | Gradient Boosting | 77.77 | 75.86 | 87.50 | **87.71** | 82.21 |
| | Bagging | 75.00 | 80.64 | 80.70 | 83.63 | 79.99 |
| | Proposed method | 77.78 | 84.38 | **90.32** | 87.27 | **84.94** |
| Accuracy | XGBoost | **85** | **85** | 85 | 85 | 85 |
| | LightGBM | 83.33 | 83.33 | 83.33 | 83.33 | 83.33 |
| | Gradient Boosting | 82.50 | 81.67 | 82.50 | 82.50 | 82.50 |
| | Bagging | 82.08 | 81.25 | 80.83 | 81.25 | 81.25 |
| | Proposed method | 80 | 83.33 | **90** | **88.33** | **85.42** |

The confusion matrix was computed for each fold and overlapped for the final model, as illustrated in Fig. 3. The confusion matrix entries acquired in all folds develop the overlapping confusion matrix. The proposed architecture correctly identified 105 healthy individuals and 100 PD patients.

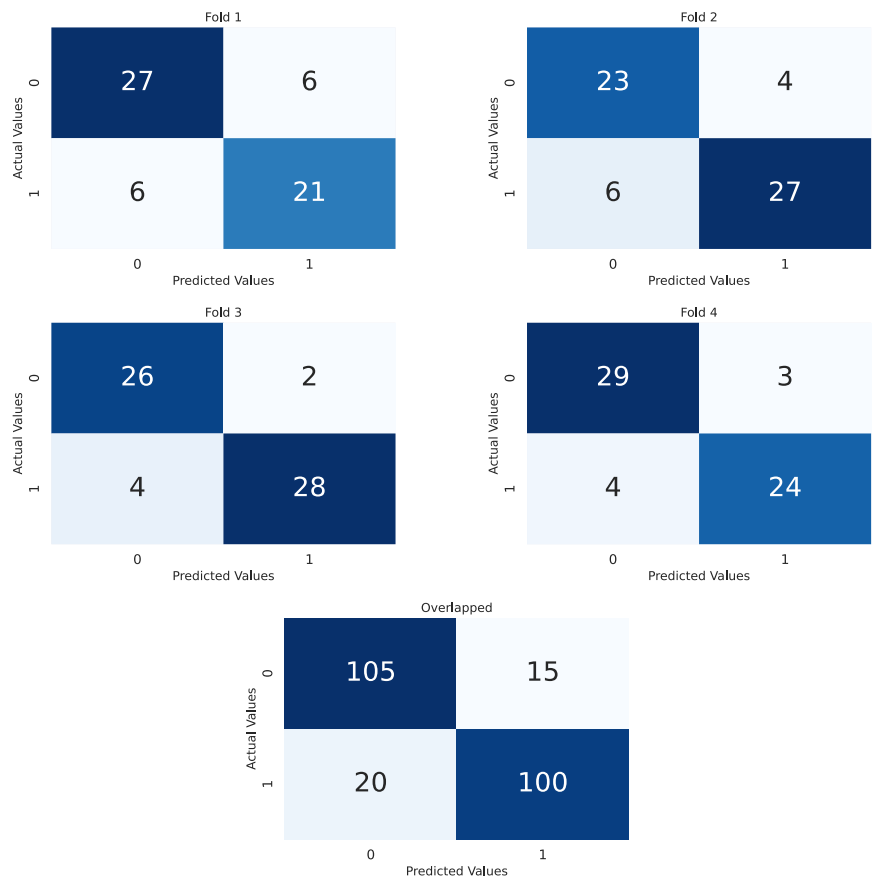

Figure 3: Confusion Matrix for proposed model

The extracted dataset was subjected to a SHAP analysis to assess the relative relevance of variables to construct a robust model. SHAP bar plot (Fig, 4) and SHAP beeswarm plot (Fig. 5) rank the acoustic features according to their relevance which is determined by the feature's mean absolute value across all samples. The fact that there are two Harmonic-to-noise ratio measurements (HNR38 and HNR35) out of five in the ranking demonstrates the significance of this feature category in diagnosing PD. Also, Glottal-to-Noise Excitation Ratio (GNE) demonstrated strong efficacy and ranked second. Moreover Delta0, MFCC9, MFCC1, MFCC10, Delta2, and Delta10 are amongst nine most important features. There is generally a strong correlation between these outcomes and Pearson Correlation assessments.

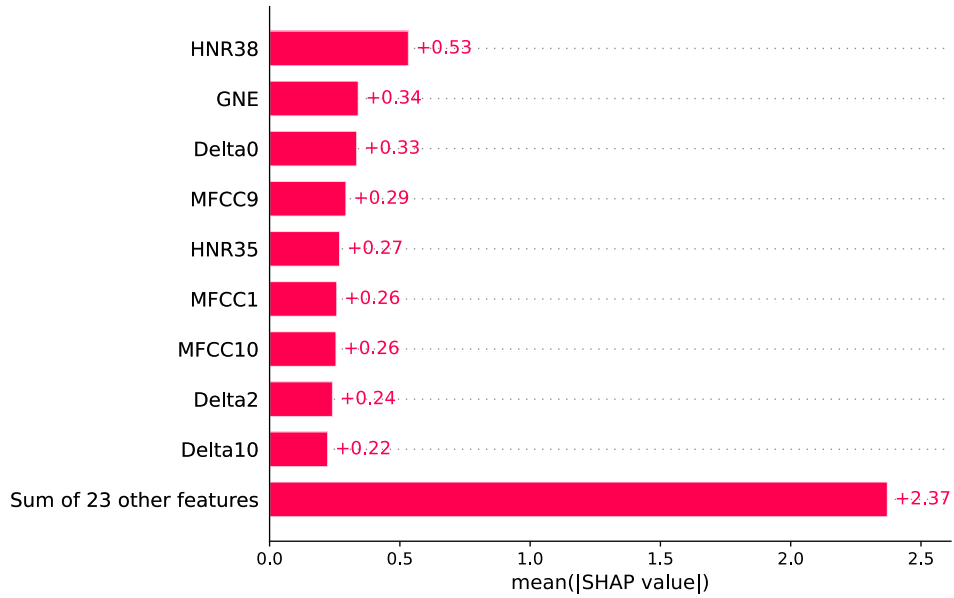

Figure 4: SHAP bar plot

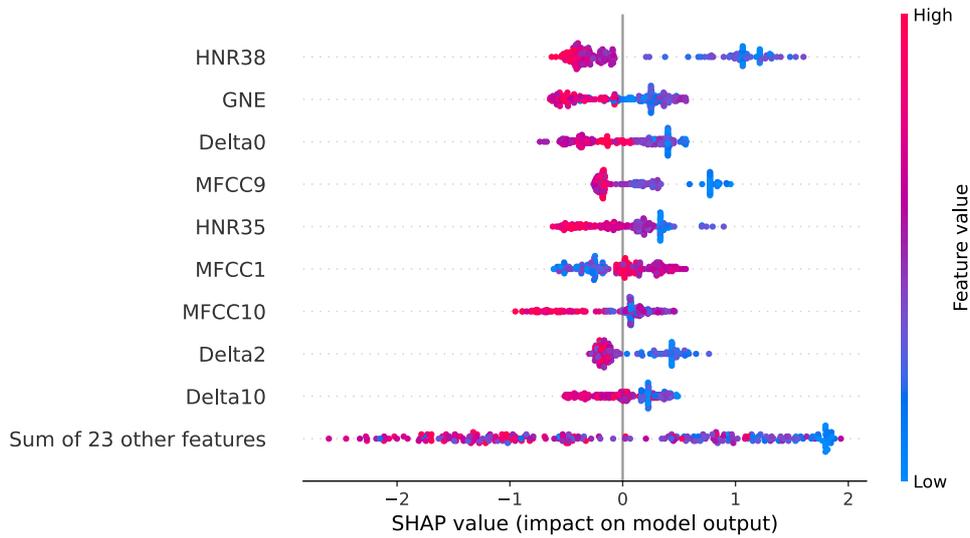

Figure 5: SHAP beeswarm plot

Table 4 summarizes the comparison results of the proposed method with Karabayir et al. [16]. Karabayir et al., achieved an 84.10% accuracy by using Parkinson Dataset with Replicated Acoustic Features. Our methodology achieved superior outcomes throughout all performance evaluation metrics except specificity. The good performance of the proposed method can be attributed to Hard Voting Ensemble Method, which integrates the beneficial characteristics of the four classifiers. The adjustability of XGBoost's parameters was a factor that contributed to its high accuracy. The high speed of LightGBM and the remarkable ability of Bagging to eliminate variance and minimize overfitting were among the factors that contributed to an increase in accuracy.

Table 4: Comparison of diagnostic validity of PD with other methods

| Study | Classifier | Specificity | Sensitivity | Precision | F$_1$-Score | Accuracy |
|---|---|---|---|---|---|---|
| Karabayir et al. [16] | XGBoost | 83.00 | 80.10 | 83.50 | 81.00 | 81.60 |
|  | Light GBM | **84.40** | 83.90 | 85.30 | 83.90 | 84.10 |
| Proposed method | Hard Voting Ensemble | 83.20 | **87.62** | **86.77** | **84.94** | **85.42** |

The Receiver Operating Characteristic (ROC) curve is one of the most important measures for evaluating the performance of a model. ROC measures classification ability at several threshold settings [56]. Therefore, we decided to compare the ROC curve of the utilized classifiers with the proposed method. In other words, it measures the power of a model to discriminate between various classes. The results are shown in Fig. 6, which also represents the area under the curve (AUC) in magnification-specific format. On the ROC curve, we can observe that our proposed model outperformed other classifiers.

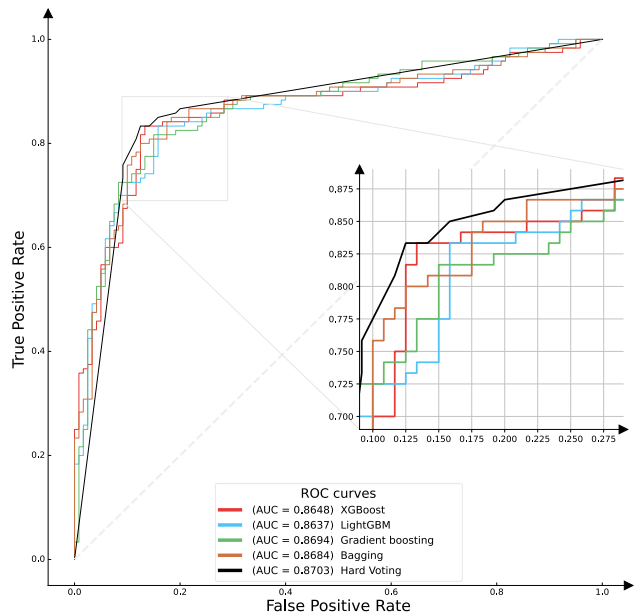

Figure 6: ROC curves of different classifiers

## 5. Conclusion

An early diagnosis approach is required since the prevalence of PD is developing. The diagnosis of PD is often made through motor symptoms or other Neuroimaging methods like DATSCAN, NMR-DTI, and SPECT. Given the constraints of these techniques, we utilized the voice recordings of people with PD, as many of them experience difficulties with their voices.

In this paper, ML algorithms play an essential role in diagnosing this disease. We initially utilized the PCCs to determine which input characteristics had a high correlation with the output. Consequently, the characteristics with the highest relation with the output were selected. In the following phase, these features were given as input to four candidate classifiers: XGBoost, LightGBM, Gradient Boosting, and Bagging. Furthermore, we used the Hard Voting Ensemble Method technique, which assigned weights to each classifier depending on its importance, and the majority vote of the classifiers determined the final prediction. The features were then ranked using SHAP to assess their relative importance. Our proposed method reached an accuracy of 85.42%, $F_1$-Score of 84.94%, precision of 86.77%, specificity of 87.62%, and sensitivity of 83.20% for Parkinson Dataset with Replicated Acoustic Features. In future work, we would like to incorporate other feature selection techniques, such as ANOVA, to achieve higher performance.

## Conflicts of Interest

The authors declare that they have no conflicts of interest.

## Funding Statement

This research received no specific grant from any funding agency in the public, commercial, or not-for-profit sectors.

## Data Availability

Publicly available Parkinson Dataset with replicated acoustic features was used in this study, which is available at UCI Machine Learning Repository[1].

## Code Availability

The source code of the proposed method required to reproduce the predictions and results is available at the public Github repository[2].

## CRediT authorship contribution statement

**Paria Ghaheri:** Conceptualization, Methodology, Software, Validation, Formal Analysis, Investigation, Writing - Original Draft, Writing - Review & Editing, Visualization. **Hamid Nasiri:** Conceptualization, Methodology, Validation, Formal Analysis, Investigation, Writing - Review & Editing, Supervision. **Ahmadreza Shateri:** Conceptualization, Methodology, Software, Formal Analysis, Investigation, Visualization. **Arman Homafar:** Methodology, Software, Formal Analysis, Investigation.

---

[1] https://archive.ics.uci.edu/ml/datasets/Parkinson+Dataset+with+replicated+acoustic+features+

[2] https://github.com/PariaGhaheri/Classification_of_Parkinson_Disease/